\documentclass[sigconf]{acmart}
\copyrightyear{2022}
\acmYear{2022}
\setcopyright{acmcopyright}\acmConference[WWW '22]{Proceedings of the ACM Web Conference 2022}{April 25--29, 2022}{Virtual Event, Lyon, France}
\acmBooktitle{Proceedings of the ACM Web Conference 2022 (WWW '22), April 25--29, 2022, Virtual Event, Lyon, France}
\acmPrice{15.00}
\acmDOI{10.1145/3485447.3512179}
\acmISBN{978-1-4503-9096-5/22/04}

\renewcommand\footnotetextcopyrightpermission[1]{} 

\usepackage{xcolor}
\usepackage[ruled,linesnumbered]{algorithm2e}
\usepackage{graphicx}
\usepackage{subfig}
\usepackage{threeparttable}
\usepackage{booktabs}
\usepackage{multirow}
\usepackage{caption}
\usepackage{url}
\usepackage{diagbox}
\usepackage{svg}
\usepackage{enumitem}
\usepackage{algorithmic}

\begin{document}
\title{Unsupervised Graph Poisoning Attack via Contrastive Loss Back-propagation}

\author{Sixiao Zhang}
\authornotemark[2]
\email{zsx57575@gmail.com}
\affiliation{%
  \institution{University of Technology Sydney}
}
\author{Hongxu Chen}
\authornote{Corresponding author.}
\authornote{Both authors contributed equally to this research.}
\email{hongxu.chen@uts.edu.au}
\affiliation{%
  \institution{University of Technology Sydney}
}
\author{Xiangguo Sun}
\email{sunxiangguo@seu.edu.cn}
\affiliation{%
  \institution{Southeast University}
}
\author{Yicong Li}
\email{Yicong.Li@student.uts.edu.au}
\affiliation{%
 \institution{University of Technology Sydney}
}
\author{Guandong Xu}
\authornotemark[1]
\email{guandong.xu@uts.edu.au}
\affiliation{%
  \institution{University of Technology Sydney}
}

\renewcommand{\shortauthors}{Zhang, et al.}

\begin{abstract}
Graph contrastive learning is the state-of-the-art unsupervised graph representation learning framework and has shown comparable performance with supervised approaches. However, evaluating whether the graph contrastive learning is robust to adversarial attacks is still an open problem because most existing graph adversarial attacks are supervised models, which means they heavily rely on labels and can only be used to evaluate the graph contrastive learning in a specific scenario. For unsupervised graph representation methods such as graph contrastive learning, it is difficult to acquire labels in real-world scenarios, making traditional supervised graph attack methods difficult to be applied to test their robustness. In this paper, we propose a novel unsupervised gradient-based adversarial attack that does not rely on labels for graph contrastive learning. We compute the gradients of the adjacency matrices of the two views and flip the edges with gradient ascent to maximize the contrastive loss. In this way, we can fully use multiple views generated by the graph contrastive learning models and pick the most informative edges without knowing their labels, and therefore can promisingly support our model adapted to more kinds of downstream tasks. Extensive experiments show that our attack outperforms unsupervised baseline attacks and has comparable performance with supervised attacks in multiple downstream tasks including node classification and link prediction. We further show that our attack can be transferred to other graph representation models as well.
\end{abstract}


\maketitle
\pagestyle{plain}
\sloppy

\section{Introduction}
Graph structured data can be commonly seen in our daily life, such as social networks \cite{ma2008sorec, zhang2021we}, biological networks \cite{zitnik2017predicting}, e-commercial networks \cite{zeng2019graphsaint}, etc. Real-world graphs usually contain rich information and can be used for various tasks including recommendation, molecular structure classification, and community detection. To extract the information contained in graphs, researchers have been exploring various machine learning methods for graph learning, including DeepWalk \cite{perozzi2014deepwalk}, node2vec \cite{grover2016node2vec}, matrix factorization \cite{rendle2012bpr}, graph neural networks \cite{scarselli2008graph} and graph convolutional networks \cite{kipf2017semi}. Graph representation learning plays an important role in these approaches, where the goal is to project nodes/graphs into a low-dimensional embedding space that preserves the structural and feature information \cite{chen2018pme, chen2019exploiting}. The quality of the embeddings directly influences the performance of downstream tasks such as link prediction, node/graph classification, community detection, etc.

 \begin{figure}[t]
\centering
    \includegraphics[width=0.45\textwidth]{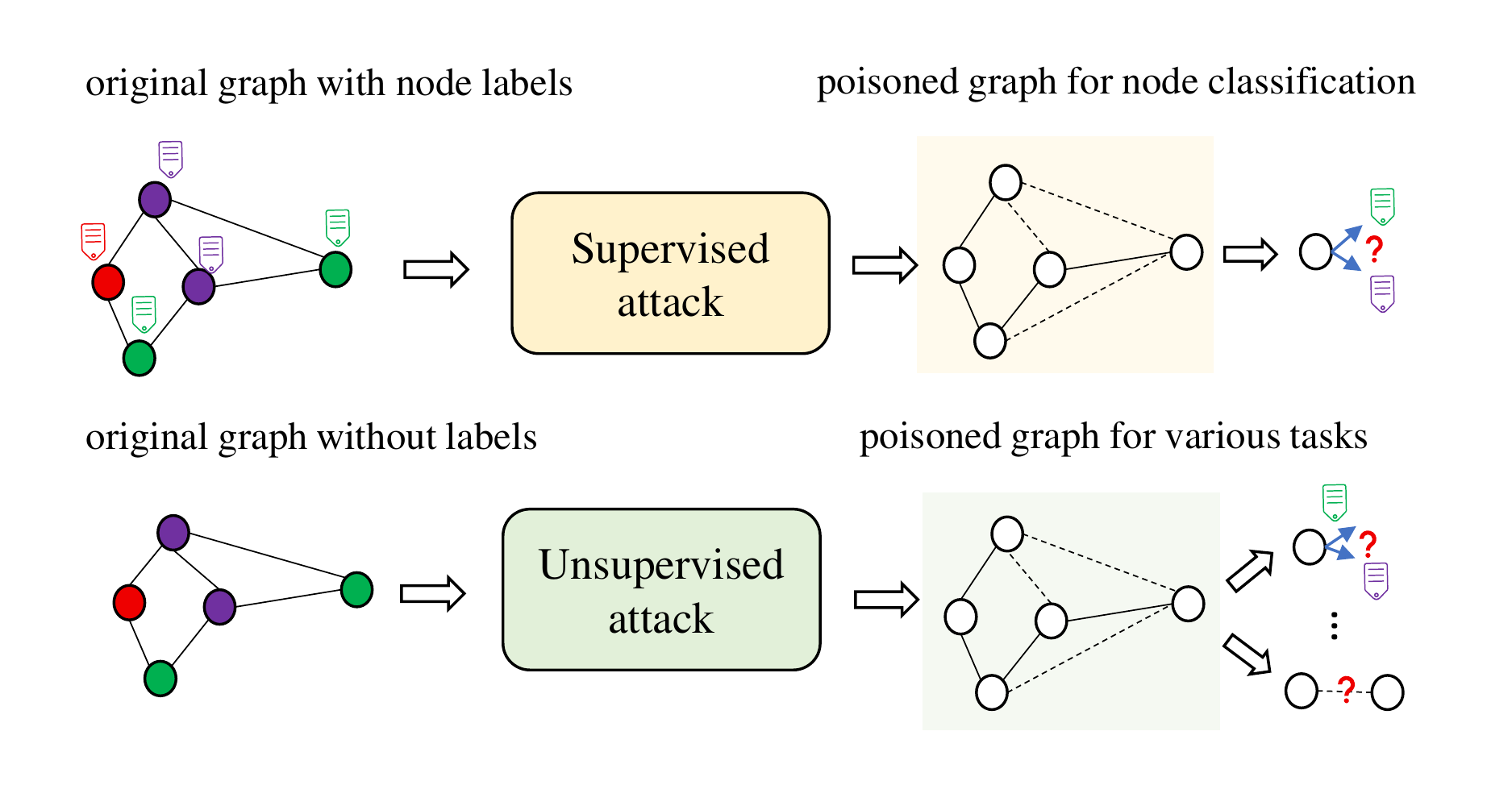}
\caption{An illustration of the difference between supervised attacks and unsupervised attacks. Supervised attacks require labels and are targeted at certain downstream tasks, whereas unsupervised attacks do not require labels and are targeted at multiple downstream tasks.}
\label{fig:supervised and unsupervised attacks}
\vspace{-1em}
\end{figure}

However, recent studies have shown that graph learning models are vulnerable to adversarial attacks \cite{jin2020adversarial}, where a small augmentation in the graph, such as adding/deleting edges/nodes and altering features, can lead to a large performance drop. The quality of the embeddings learned by graph representation models is also sensitive to such attacks \cite{chang2019general, zhang2019towards}, and will in turn affects the performance of the downstream tasks. In addition, adversarial attacks in the graph domain are easy to conduct. For example, the attacker can attack a social network by simply creating some fake users and establishing connections with other people by following their accounts. The e-commercial networks can also be attacked by creating fake user profiles and writing fake reviews. Therefore, it is important to study the robustness of various graph learning models and develop models that are robust to adversarial attacks.

Recently, researchers have been exploring the robustness of supervised and semi-supervised graph representation models \cite{dai2018adversarial, zugner2018adversarial, zhu2019robust, jin2020graph, zhang2021graph}, but the robustness of unsupervised graph representation models still remains an open challenge. Due to the complexity and scale-freeness of many real-world graphs, the labels are hard to acquire in most cases \cite{velivckovic2019deep}. It is also impractical to do manual annotation for unlabeled large graphs. Thus, unsupervised graph representation learning becomes an important branch in graph representation learning. However, most state-of-the-art graph representation learning methods are supervised or semi-supervised \cite{kipf2017semi, velivckovic2018graph}. Traditional unsupervised methods, such as DeepWalk \cite{perozzi2014deepwalk} and node2vec \cite{grover2016node2vec}, have a relatively lower performance on downstream tasks compared with supervised methods. However, unsupervised graph representation learning has again attracted the attention of researchers in recent years because of the success of contrastive learning \cite{you2020graph, zhu2020graph, qiu2020gcc, zhu2020deep, yang2021hyper}. It has emerged as a new state-of-the-art unsupervised learning framework and has shown comparable performance with supervised and semi-supervised baselines. It generates different views of the original graph using stochastic augmentations and adopts a special contrastive loss to learn embeddings by comparing these views, which also makes graph contrastive learning models more robust to adversarial attacks compared with other graph representation methods \cite{you2020graph}. However, such conclusions are obtained using existing graph adversarial attacks, which are mostly supervised attacks \cite{xu2019topology, dai2018adversarial, zugner2018adversarial, zugner2019adversarial}, i.e. they need labels to conduct the attack as shown in \autoref{fig:supervised and unsupervised attacks}. In real-world scenarios, it is difficult to attack contrastive learning with supervised attacks because the labels are hard to acquire. Therefore, this raises a new challenge that how to attack unsupervised graph representation learning such as contrastive learning without knowing the labels? Previous work by Bojchevski et al. \cite{bojchevski2019adversarial} has shown that unsupervised graph learning methods based on random walks such as DeepWalk can be attacked in an unsupervised way by approximating the influence of a single augmentation in an optimization perspective. However, their method is limited to random walk and we will show in our experiments that it can not perform well on graph contrastive learning.


Attacking graph contrastive learning in an unsupervised way is a non-trivial task. Our goal is to poison the graph so that the overall performance on various downstream tasks is degraded. But we can't use the downstream task to measure the quality of the embeddings. Therefore, how to explicitly measure the quality of the embeddings is the first problem needed to be solved. In addition, the stochastic augmentation process provides robustness against adversarial attacks, so the attack needs to be insensitive to stochastic augmentations. Besides, a powerful graph attack model should let the poisoned graph be deterrent as much as possible, which means our generated poisoned graphs should be also transferable to the other graph representation models.

To solve the above challenges, we propose a novel unsupervised adversarial attack based on gradient ascent on graph contrastive learning for node embeddings, which is called \textbf{C}ontrastive \textbf{L}oss \textbf{G}radient \textbf{A}ttack (CLGA). To the best of our knowledge, this is the first work aiming at attacking graph contrastive learning in an unsupervised manner. Specifically, we compute the gradient of the contrastive loss w.r.t. the adjacency matrix, and flip the edges with the largest gradients, causing the contrastive loss to be damaged. We show by extensive experiments that CLGA outperforms other existing unsupervised attack baselines and has comparable performance with some of the supervised attack methods. We also show that CLGA can be transferred to other graph representation models. To guarantee reproducibility, we open the source code at \footnote{\url{https://github.com/RinneSz/CLGA}}. In summary, our contributions are as follows:
\begin{itemize}
    \item We propose CLGA, a gradient-based unsupervised attack method targeting graph contrastive learning. Unlike most supervised attack models, CLGA does not rely on labels and can degrade the quality of the learned embeddings and thus affect the performance of various downstream tasks.
    \item We show by extensive experiments that CLGA outperforms unsupervised attack baselines and has comparable performance with some of the supervised attack methods on three benchmark datasets and on both node classification and link prediction tasks.
    \item We also show that CLGA can be transferred to other graph representation models such as GCN and DeepWalk.
    \item We visualize the learned embeddings to show how CLGA influences the quality of them.
\end{itemize}
In the following content, we first introduce the related work and preliminaries for graph adversarial attack and graph contrastive learning. Next we introduce how CLGA works. Moreover, we show the experiment results on three benchmark datasets and compare CLGA with several state-of-the-art baseline attacks. At last, we provide the conclusion of the paper.
\section{Related Work}
\subsection{Graph Adversarial Attack}
The robustness of various machine learning models is a hot topic in the research community. A small perturbation in the training data might lead to a big change in the model performance. How to attack models and how to defend against such attacks are two main streams of this field. In the graph domain, people have already proposed many successful adversarial attack methods. RL-S2V \cite{dai2018adversarial} adopts reinforcement learning to attack graph neural networks by adding/deleting edges that have a positive effect on the attacker's goal. Later, NIPA \cite{sun2019node} and ReWatt \cite{ma2019attacking} also use reinforcement learning to attack GNNs but only by injecting nodes and rewiring edges respectively. Some works try to model the attack as an optimization problem. PGD and MinMax \cite{xu2019topology} optimize the negative cross-entropy loss using the gradient. Nettack \cite{zugner2018adversarial} iteratively selects augmentations by calculating the score/loss of each possible augmentation and chooses the one that maximizes the loss. Besides, because of the strong ability of gradients to locate the most important instances, gradient-based attack methods have emerged as new state-of-the-arts, such as Mettack \cite{zugner2019adversarial} and FGA \cite{chen2018fast}, where they both use the gradient of the classification loss w.r.t. the adjacency matrix to select the edges to augment. However, all the attack methods mentioned above are supervised attack methods, which means that they require labels to conduct the attack, either by using the ground-truth labels or by using the predictions of the target model as labels. They cannot be directly used to attack an unsupervised model that aims at learning pre-trained embeddings. Bojchevski et al. \cite{bojchevski2019adversarial} have proposed an unsupervised optimization-based attack method to attack graph embedding methods based on random walks such as DeepWalk without using the labels. But we will show that it does not work very well for the state-of-the-art contrastive learning framework.

\subsection{Graph Contrastive Learning}
Recently, because of the great success of contrastive learning in computer vision \cite{he2020momentum, chen2020simple}, researchers have begun to explore contrastive learning in the graph domain and have achieved competitive performance compared with supervised and semi-supervised models. DGI \cite{velivckovic2019deep} maximizes the mutual information between patch representations and corresponding high-level summaries of graphs. CSSL \cite{zeng2020contrastive} first defines four types of basic augmentation operations, edge deletion, node deletion, edge insertion and node insertion, wherein all of them choose nodes/edges randomly. Then they randomly sample a sequence of operations to apply to the graph. GraphCL \cite{you2020graph} maximizes the agreement between two views by uniform perturbations including node dropping, edge perturbation, attribute masking and subgraphs. GCC \cite{qiu2020gcc} uses random walk on the r-ego network to generate different subgraphs, and learns embeddings by comparing these subgraphs. GRACE \cite{zhu2020deep} learns node embeddings by randomly removing edges and masking features. GCA \cite{zhu2020graph} further improves GRACE by designing new augmentation schemes that are aimed to keep important structures and attributes unchanged, while perturbing possibly unimportant links and features. These works differ in the specific augmentation strategy, such as adding/deleting edges/nodes, masking features, and extracting subgraphs. The loss functions used in these works share the same idea. They treat the embeddings of the same instance in different augmented views as positive pairs and treat the embeddings of different instances as negative pairs. Later, bootstrapped graph contrastive learning models have been proposed by Thakoor et al \cite{thakoor2021bootstrapped} and Che et al \cite{che2020self}, which do not require negative pairs. Such a method refers to two neural networks, the online network and the target network. Two augmented views are the input to the two networks respectively. The online network is trained to predict the target network output, and the target network is updated by an exponential moving average of the online network. After training, the encoder of the online network is used to compute the representations of the downstream tasks.

\section{Preliminaries}
\subsection{Graph Contrastive Learning on Node Embeddings}
A graph $G$ is defined as $G=(V,E)$, where $V$ is the set of vertices (nodes) and $E$ is the set of edges. It can also be represented by the adjacency matrix $A\in\mathbb{R}^{N\times N}$, where $N$ is the number of nodes. Some graphs also have features $X\in\mathbb{R}^{N\times d}$ associated with nodes, where $d$ is the number of features. The goal of graph contrastive learning is to learn an encoder $f(A,X)$ that outputs embeddings of each node. Such embeddings can then be used for downstream tasks including node classification and link prediction.

In this paper, we focus on node-level contrastive learning. So we will introduce how node-level graph contrastive learning works following the state-of-the-art GCA model \cite{zhu2020graph}. A general graph contrastive learning framework consists of three steps. First, two stochastic augmentation processes $t_{1}$ and $t_{2}$ are applied to the original graph to obtain two different views. Typical augmentation strategies include edge dropping/insertion, feature masking, subgraph extracting, etc. Second, the two views are fed into a shared encoder $f(A,X)$ to obtain node embeddings. Each node will have two embeddings corresponding to the two views. Third, a contrastive loss is applied to the embeddings to push positive node pairs close to each other and push negative node pairs away from each other. Specifically, the loss for the $i$-th node in the first view is
\begin{equation}
    l(e_{i}^{1},e_{i}^{2})=-\text{log}\frac{e^{\beta(e_{i}^{1},e_{i}^{2})/\tau}}{e^{\beta(e_{i}^{1},e_{i}^{2})/\tau}+\sum_{j\neq i}(e^{\beta(e_{i}^{1},e_{j}^{1})/\tau}+e^{\beta(e_{i}^{1},e_{j}^{2})/\tau})}
\end{equation}
where $e_{i}^{1}$ and $e_{i}^{2}$ denote the embeddings of the $i$-th node in the first and second view respectively. $\beta$ is a similarity function, e.g. cosine similarity. $\tau$ is a temperature parameter. Note that the above loss is non-symmetric for $e_{i}^{1}$ and $e_{i}^{2}$. So we need to add it up with its counterpart $l(e_{i}^{2},e_{i}^{1})$. The final loss $\mathcal{L}$ is the sum of $l$ for all nodes, which is
\begin{equation}
\label{eq:final loss}
    \mathcal{L}=\sum_{i=1}^{N}l(e_{i}^{1},e_{i}^{2})+l(e_{i}^{2},e_{i}^{1})
\end{equation}
By minimizing $\mathcal{L}$, the model treats the embeddings of the same node in the two views as positive pairs and the embeddings of different nodes in either the same view or different views as negative pairs. Because of the stochastic augmentation procedure, the learned embeddings are more robust to the small perturbations in the graph compared with other conventional graph representation models.

\begin{figure*}[h]
\centering
    \includegraphics[width=0.8\textwidth]{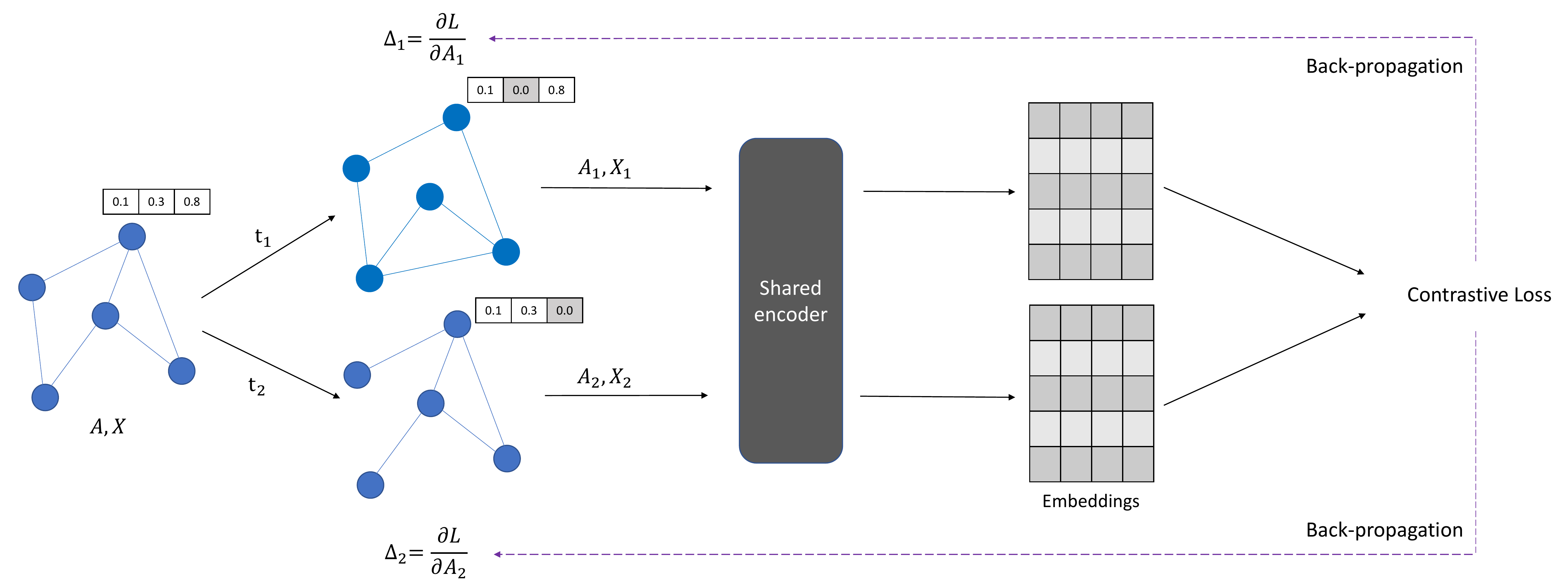}
\vspace{-1.3em}
\caption{CLGA framework. The clean adjacency matrix $A$ and feature matrix $X$ are augmented by two stochastic augmentations $t_{1}$ and $t_{2}$ to obtain two views $A_{1},X_{1}$ and $A_{2},X_{2}$. The two views are then fed into a shared encoder to obtain the node embeddings. The contrastive loss is computed  using the embeddings. We then back-propagate the contrastive loss to obtain the gradients of the two views' adjacency matrices. The gradients are used for selecting the edges that will be flipped in our attack.}
\label{fig:CLGA}
\vspace{-1em}
\end{figure*}

\subsection{Graph Adversarial Attack}
Adversarial attack aims to add noise to the data to degrade the performance of the target model. For graph data, the common noise includes edge-level, node-level, and graph-level augmentations, such as adding/deleting edges/nodes, augmenting features, and adding fake graphs for graph classification. 

According to the goal of the attacker, adversarial attacks can be divided into two categories:
\begin{itemize}
    \item \textbf{Untargeted Attack}: force the model to have bad overall performance on all given instances.
    \item \textbf{Targeted Attack}: force the target model to have bad performance on a subset of instances.
\end{itemize}
A successful untargeted attack will make the target model produce a biased prediction for every query, and a targeted attack will make the target model produce a biased prediction for only certain queries. As a result, when testing the performance of the model on a test set, a successful untargeted attack will make the model have a worse overall accuracy compared with the model trained on clean data. A successful targeted attack will only make the model produce wrong predictions on a specific subset of the test set, while still producing correct predictions for other data, and thus the overall accuracy may not be significantly reduced.

Besides, according to the capacity of the attacker, adversarial attacks can be divided into two categories:
\begin{itemize}
    \item \textbf{Poisoning Attack}: attack happens before the target model is trained. The training data of the target model is poisoned and is being used to train the target model.
    \item \textbf{Evasion Attack}: attack happens after the model is trained. The model is trained on clean training data and is fixed.
\end{itemize}
Poisoning attacks are a stronger attack since the attacker can affect the learning process of the target model. For evasion attacks, since the target model is fixed, what we can do is to perturb the data to cause misclassifications. For example, the attacker can modify the graph of a protein molecule to make it be misclassified as another type of protein.

For node representation learning on graphs, evasion attack is hard to conduct due to the transductive setting where all nodes are seen by the target model during training. Therefore, in this paper, we focus on the untargeted poisoning attack since this is best suited for attacking node representation learning models. Targeted poisoning attacks are a special case for untargeted poisoning attacks and can be achieved by restricting the target nodes.

\subsection{Graph Convolutional Networks}
The graph convolutional network proposed by Kipf et al. \cite{kipf2017semi} is usually used as the encoder $f$ in graph contrastive learning because of its state-of-the-art performance in modeling graph-structured data. Given the adjacency matrix $A$ and feature matrix $X$, the output $X'$ of a typical GCN layer is computed by
\begin{equation}
\label{eq:GCN}
    X'=\widetilde{D}^{-1/2}\widetilde{A}\widetilde{D}^{-1/2}X\Theta
\end{equation}
where $\widetilde{A}=A+I$ denotes the adjacency matrix with self-loops. $\widetilde{D}$ is a diagonal degree matrix and $\widetilde{D}_{ii}=\sum_{j}\widetilde{A}_{ij}$. $\Theta$ is the parameter matrix of the GCN layer. Activation functions such as ReLU and Sigmoid are applied at the output of each layer. In conventional GCN models, $\Theta$ is the only learnable parameter in \autoref{eq:GCN} and is updated by gradient descent. However, to poison the graph, we need to know the gradient on the adjacency matrix $A$ and luckily it is also easy to be deduced by \autoref{eq:GCN}. We will introduce how we attack graph contrastive learning using the gradient of the adjacency matrix in the next section.
\section{Contrastive Loss Gradient Attack}
\subsection{Overview}
We want to design an untargeted poisoning attack, where the goal is to poison the graph data such that the overall quality of the embeddings learned by graph contrastive learning is degraded, which leads to worse performance in multiple downstream tasks. The contrastive loss used in contrastive learning is a natural measurement of the embedding quality. Therefore, we choose to poison the graph by maximizing the contrastive loss. In our attack, we only augment edges and do not augment features, because features are auxiliary information and not all the graphs have features. We formulate our problem as:
\begin{equation}
\label{eq:formulation}
\begin{split}
    \max\limits_{\hat{A}}\quad&\mathcal{L}(f_{\theta'}(A_{1},X_{1}),f_{\theta'}(A_{2},X_{2}))\\
    \text{s.t.}\quad&\theta'=\mathop{\arg\min}\limits_{\theta} \mathcal{L}(f_{\theta}(A_{1},X_{1}),f_{\theta}(A_{2},X_{2})),\\
    &(A_{1},X_{1})=t_{1}(\hat{A},X), (A_{2},X_{2})=t_{2}(\hat{A},X),\|A-\hat{A}\|=\sigma.\\
\end{split}
\end{equation}
here $\mathcal{L}$ is the contrastive loss stated in \autoref{eq:final loss}. $f$ is the encoder and $\theta$ is the set of parameters of the encoder. $A$ and $X$ are the adjacency matrix and feature matrix of the clean graph. $\hat{A}$ is the poisoned adjacency matrix. $t_{1}$ and $t_{2}$ are two stochastic augmentations. $A_{1}$ and $X_{1}$ are the adjacency matrix and feature matrix of the first view, and $A_{2}$ and $X_{2}$ are those of the second view. The last constraint indicates that the number of augmented edges is bounded by a given threshold $\sigma$.

This is a bi-level optimization problem and is hard to be solved directly. Inspired by \cite{zugner2019adversarial}, we propose to use meta-gradients, i.e. gradients w.r.t. the adjacency matrix. We back-propagate the contrastive loss to obtain the gradients of the adjacency matrix and update the adjacency matrix to maximize the loss.

\subsection{Gradient based Attack}
To solve \autoref{eq:formulation}, we propose to use gradient ascent on the adjacency matrix. The key idea is to flip the edges with the largest gradient values and the correct gradient directions, and thus the contrastive loss will be maximized. For example, if an observed edge (which means the corresponding entry is 1 in the adjacency matrix) has a negative gradient, then deleting it (from 1 to 0 in the adjacency matrix) will be very likely to increase the loss. Similarly, for an unobserved edge that has a large positive gradient, adding it to the graph will also increase the loss. An illustration of how we acquire the gradients is shown in \autoref{fig:CLGA}. Specifically, if we use a differentiable encoder $f(A,X)$, e.g. graph convolutional network (GCN), we can easily obtain the gradients of the two views' adjacency matrix $A_{1}$ and $A_{2}$:
\begin{equation}
    \Delta_{1}=\frac{\partial \mathcal{L}}{\partial A_{1}}=\frac{\partial \mathcal{L}}{\partial f(A_{1},X_{1})}\cdot\frac{\partial f(A_{1},X_{1})}{\partial A_{1}}
\end{equation}
\begin{equation}
    \Delta_{2}=\frac{\partial \mathcal{L}}{\partial A_{2}}=\frac{\partial \mathcal{L}}{\partial f(A_{2},X_{2})}\cdot\frac{\partial f(A_{2},X_{2})}{\partial A_{2}}
\end{equation}
Ideally, the most informative edges usually contribute largely to the graph learning model because their loss gradients on the clean adjacency matrix $\Delta=\frac{\partial \mathcal{L}}{\partial A}$ usually have larger absolute values. To find these edges, we need to differentiate through the stochastic augmentations $t_{1}$ and $t_{2}$ to obtain $\Delta$ as shown in \autoref{fig:CLGA}. Unfortunately, the stochastic augmentation process $t$ might contain some indifferentiable policy such as adding/deleting nodes and extracting subgraphs, making the problem even harder. To solve this problem, we do not back-propagate through $t$ and only use $\Delta_{1}$ and $\Delta_{2}$ to help select which edges to flip. Another problem is that we cannot directly use $\Delta_{1}$ or $\Delta_{2}$ to select edges because the adjacency matrices of the two views are different from the clean adjacency matrix due to the stochastic augmentations. That means the edges having the largest gradients in these two views ($\Delta_{1}$ and $\Delta_{2}$) might not be the truly important ones in the original graph. With the above discussion, we can find that the core challenge is how to combine these two gradient matrices $\Delta_{1}$ and $\Delta_{2}$ to find the most informative edges and in the meanwhile, alleviate the bias caused by the stochastic augmentations ($t_{1}$ and $t_{2}$) as much as possible.

But what exactly is the form of the bias caused by the stochastic augmentations? We answer this question with a small example. As we know, conventional graph contrastive learning learns embeddings that are insensitive to the change of the graph by updating the parameters of the encoder. It tries to eliminate the difference between the two views. If we assume that the adjacency matrices of the two views are learnable parameters, how will the adjacency matrices be like? Apparently they will become identical. Consider that if the two views are already identical, then every edge will have a zero gradient. If we now apply a stochastic augmentation and one single edge is changed, where it exists in the first view but does not exist in the second view, it will have a negative gradient in the first view and a positive gradient in the second view, pushing the corresponding entries 1 and 0 closer to each other. In this case, this augmented edge will have a much larger gradient compared with other edges, but can we say that it is more informative than other edges? Of course not, because the large gradient is actually caused by the stochastic augmentation. Instead, if we add the two gradients up, where one is positive and another is negative, the result will be close to 0 and we will thus acquire a relatively accurate measurement of how important this edge is by successfully alleviating the bias introduced by the stochastic augmentation.

With the above example, we propose to solve the problem by using two little tricks. First, we add $\Delta_{1}$ and $\Delta_{2}$ up to alleviate the bias introduced by the stochastic augmentations:
\begin{equation}
\Delta_{0}=\Delta_{1}+\Delta_{2}
\end{equation}
In addition, we add up gradients of $K$ random stochastic augmentation pairs to further alleviate the bias caused by some rare cases:
\begin{equation}
\label{eq:k iterations}
    \Delta'=\sum_{k=1}^{K}\Delta_{0}^{k}=\Delta_{1}^{k}+\Delta_{2}^{k}
\end{equation}
where for each $k$, $\Delta_{1}^{k}$ and $\Delta_{2}^{k}$ is obtained by a random stochastic augmentation pair $t_{1}^{k}$ and $t_{2}^{k}$. We use $\Delta'$ to select the edges that have the largest absolute gradients and correct gradient directions to flip.

Specifically, if for an edge, the gradients in the two views are both positive, it suggests that even if the stochastic augmentations are applied, the two views both prefer a smaller value in the corresponding entry which will lead to a smaller contrastive loss. Instead, increasing the value in this entry by adding this edge (if it does not exist) is very likely to increase the contrastive loss. This also applies to the case where the gradients are both negative. In our method, we add the two gradients to obtain a larger absolute value and this edge will be ranked at a higher position. If one of the gradients is positive and another is negative, it is usually caused by the stochastic augmentation on itself or its neighborhood. In this case, the information contained in its gradients is more about how to compensate for the heavily augmented neighborhood, or how to alleviate the changes caused by the stochastic augmentations. We are unable to tell how important this edge itself is w.r.t. the contrastive loss, so we hope that this edge will be ranked in a lower place when we are picking which edges to flip. We can achieve this by adding the two gradients up to get a new value that has a smaller absolute value than before, and it will thus have a lower ranking when we rank all the edges. 

To further improve the attack, in each iteration we only pick one edge to flip. Specifically, in the first iteration, we train the contrastive model with the clean graph. Then we compute $\Delta'$ and select one edge with the largest gradient and correct direction. We flip the selected edge and use the new adjacency matrix (which differs from the clean adjacency only on this single edge) to retrain the contrastive model in the next iteration. So if we are going to flip 100 edges, we will need to retrain the model 100 times. This iterative process helps us to better locate informative edges.

The overall algorithm is shown in \autoref{alg:CLGA}.

\begin{algorithm}[h]
\caption{CLGA}
\label{alg:CLGA}
\begin{algorithmic}[1]
\REQUIRE
Clean adjacency matrix $A$, feature matrix $X$, differentiable encoder $f$, stochastic augmentation set $T$, augmentation threshold $\sigma$, number of iterations $K$;
\ENSURE
Poisoned graph $\hat{A}$;
\STATE $i = 0; \hat{A}=A$;
\WHILE{$i < \sigma$}
\STATE Train $f$ with $\hat{A}$ and $X$;
\STATE $\Delta'=0$;
\FOR{$k=1$ to $K$}
\STATE Sample two stochastic augmentations $t_{1}^{k},t_{2}^{k}\in T$;
\STATE Obtain two views $(A_{1}^{k},X_{1}^{k})=t_{1}^{k}(\hat{A}, X)$, $(A_{2}^{k},X_{2}^{k})=t_{2}^{k}(\hat{A}, X)$;
\STATE Forward propagate $(A_{1}^{k},X_{1}^{k})$, $(A_{2}^{k},X_{2}^{k})$ through $f$ and compute contrastive loss $\mathcal{L}$;
\STATE Obtain the gradients of $A_{1}^{k}$ and $A_{2}^{k}$ w.r.t. the contrastive loss, $\Delta_{1}^{k}=\frac{\partial\mathcal{L}}{\partial A_{1}^{k}}$, $\Delta_{2}^{k}=\frac{\partial\mathcal{L}}{\partial A_{2}^{k}}$;
\STATE $\Delta'=\Delta'+\Delta_{1}^{k}+\Delta_{2}^{k}$;
\ENDFOR
\STATE Flip one edge with both the largest absolute gradient in $\Delta'$ and the correct direction, i.e., if the index of the edge is $[m,n]$, then it should satisfy either $\hat{A}[m,n]=1$, $\Delta'[m,n]<0$ or $\hat{A}[m,n]=0$, $\Delta'[m,n]>0$;
\STATE $\hat{A}[m,n]=1-\hat{A}[m,n]$;
\STATE Freeze the chosen edge and avoid being flipped again in next iterations;
\STATE $i = i + 1$;
\ENDWHILE
\end{algorithmic}
\end{algorithm}

\subsection{Complexity Analysis}
\subsubsection{Time Complexity}
The time complexity of CLGA itself is low since we only need to rank the gradients and select the desired edges. What is truly time-consuming is the training of the target model, because we need to retrain it in each iteration. Therefore, for complex models whose time complexity is large, it will take a relatively long time for the poisoned graph to be generated. However, we can alleviate this issue by doing some approximations at the price of attack performance. For example, we can select multiple edges in each iteration, or we can train the target model with pre-training or early-stopping because we do not necessarily need the target model to be fully converged as long as the order of the ranking is correct.

\subsubsection{Space Complexity}
Since we are computing the gradients of the whole adjacency matrix, the memory requirement is $O(N^{2})$, where $N$ is the number of nodes. This is expensive for graphs with a large number of nodes. However, we can also alleviate this issue by only considering a subset of nodes and edges. For real-world scale-free graphs, it is impractical for the attackers to be able to modify the whole graph, where they can only manipulate a small subset of nodes and edges. In this case, we only need to compute the gradients of the subset, which only has a small memory cost.

\section{Experiments}
In our experiments, we aim to answer the following research questions:
\begin{itemize}
    \item Can CLGA successfully degrade the performance of various downstream tasks of graph contrastive learning?
    \item Does CLGA outperform other unsupervised and supervised attacks?
    \item Is the poisoned graph generated by CLGA able to degrade the performance of other graph representation models?
\end{itemize}
To answer these questions, we compare CLGA with five state-of-the-art representative graph untargeted poisoning attacks on three benchmark datasets. We evaluate the quality of the embeddings by two downstream tasks: node classification and link prediction. We further show the transferability of CLGA by evaluating other graph representation models on the obtained poisoned graph.

\subsection{Setup}
\subsubsection{Datasets}
We use three popular public available benchmark datasets. The two of them are citation networks Cora and CiteSeer from Yang et al. \cite{yang2016revisiting}. We also use PolBlog dataset from Adamic et al. \cite{adamic2005political}, which is a graph of political blogs. There are features associated with each node in Cora and CiteSeer. However, there are no features for PolBlog. The statistics of the three datasets are shown in \autoref{table:datasets}.
\begin{table}[t]
\begin{center}
 \begin{tabular}{|c|c|c|c|c|} 
 \hline
 Dataset & \# Nodes & \# Edges & \# Features & \# Classes\\
 \hline
 Cora & 2708 & 5278 & 1433 & 7 \\
 \hline
 CiteSeer & 3327 & 4552 & 3703 & 6 \\
 \hline
 PolBlogs & 1490 & 16715 & None & 2 \\
 \hline
\end{tabular}
\caption{Dataset statistics.}
\label{table:datasets}
\end{center}
\vspace{-2em}
\end{table}

\begin{table*}
\begin{center}
 \begin{tabular}{c|c||c|c|c||c|c|c||c|c|c} 
 \multirow{2}{*}{} & \multirow{2}{*}{Attack} & \multicolumn{3}{c||}{Cora} & \multicolumn{3}{c||}{CiteSeer}& \multicolumn{3}{c}{PolBlog}\\
 \cline{3-11}
  & & 1\% & 5\% & 10\% & 1\% & 5\% & 10\% & 1\% & 5\% & 10\%\\
 \hline
 \multirow{3}{*}{Supervised} & Metattack & \textbf{0.7586} & \textbf{0.6928} & \textbf{0.6168} & \textbf{0.5920} & \textbf{0.3986} & \textbf{0.2952} & 0.8208 & 0.8039 & 0.8011\\
 \cline{2-11}
 &PGD & 0.7680 & 0.7592 & 0.7402 & 0.6098 & 0.6198 & 0.6056 & 0.8100 & 0.8010 & 0.7987\\
 \cline{2-11}
 &MinMax & 0.7624 & 0.7218 & 0.6174 & 0.6302 & 0.5254 & 0.5618 & \textbf{0.8016} & 0.7913 & 0.7986\\
 \cline{2-11}
 &DICE & 0.7712 & 0.7642 & 0.7240 & 0.6256 & 0.5774 & 0.5246 & 0.8107 & \textbf{0.7847} & \textbf{0.7394}\\
 \hline
 \hline
 \multirow{2}{*}{Unsupervised}& Bojchevski et al. \cite{bojchevski2019adversarial} & 0.7490 & 0.7710 & 0.7670 & 0.6442 & 0.6448 & 0.6608 & 0.8187 & 0.8042 & 0.7892\\
 \cline{2-11}
 & \textbf{CLGA} & \textbf{0.7316} & \textbf{0.7188} & \textbf{0.6814} & \textbf{0.6368} & \textbf{0.5906} & \textbf{0.5368} & \textbf{0.8088} & \textbf{0.7944} & \textbf{0.7726}\\
\end{tabular}
\caption{Node classification accuracy of logistic regression model trained after GCA. 1\%/5\%/10\% denote the maximum number of edges allowed to be augmented. The boldfaced ones are the best in either supervised or unsupervised approaches.}
\label{table:node classification performance}
\end{center}
\vspace{-2em}
\end{table*}

\begin{table*}
\begin{center}
 \begin{tabular}{c|c||c|c|c||c|c|c||c|c|c} 
 \multirow{2}{*}{} & \multirow{2}{*}{Attack} & \multicolumn{3}{c||}{Cora} & \multicolumn{3}{c||}{CiteSeer}& \multicolumn{3}{c}{PolBlog}\\
 \cline{3-11}
  & & 1\% & 5\% & 10\% & 1\% & 5\% & 10\% & 1\% & 5\% & 10\%\\
 \hline
 \multirow{3}{*}{Supervised} & Metattack & \textbf{0.9010} & \textbf{0.8733} & \underline{0.8500} & \textbf{0.9109} & \textbf{0.8853} & \textbf{0.8544} & 0.8617 & 0.8585 & 0.8635\\
 \cline{2-11}
 &PGD & 0.9143 & 0.9073 & 0.9073 & 0.9169 & 0.9248 & 0.9057 & 0.8605 & 0.8584 & 0.8625\\
 \cline{2-11}
 &MinMax & 0.9116 & 0.9004 & 0.8944 & 0.9145 & \underline{0.8890} & 0.8981 & 0.9145 & 0.8890 & 0.8981\\
 \cline{2-11}
 &DICE & 0.9046 & 0.8828 & 0.8593 & 0.9137 & 0.8918 & 0.8679 & \textbf{0.8551} & \textbf{0.8450} & \textbf{0.8352}\\
 \hline
 \hline
 \multirow{2}{*}{Unsupervised}& Bojchevski et al. \cite{bojchevski2019adversarial} & 0.9164 & 0.9099 & 0.9101 & 0.9239 & 0.9168 & 0.9196 & 0.8593 & \underline{0.8543} & 0.8587\\
 \cline{2-11}
 & \textbf{CLGA} & \underline{0.9012} & \underline{0.8741} & \textbf{0.8420} & \underline{0.9114} & 0.8911 & \underline{0.8610} & \underline{0.8584} & 0.8598 & \underline{0.8563}\\
\end{tabular}
\caption{Link prediction AUC of the MLP trained after GCA. 1\%/5\%/10\% denote the maximum number of edges allowed to be augmented. The boldfaced ones are the best and the underlined ones are the second best among all approaches.}
\label{table:link prediction performance}
\end{center}
\vspace{-2em}
\end{table*}

\subsubsection{Baselines}
We compare CLGA with five baseline untargeted poisoning attacks, including PGD \cite{xu2019topology}, DICE \cite{waniek2018hiding}, MinMax \cite{xu2019topology}, Metattack \cite{zugner2019adversarial}, and the unsupervised node embedding attack proposed by Bojchevski et al. \cite{bojchevski2019adversarial}. 
Among these baselines, only Bojchevski et al. \cite{bojchevski2019adversarial} is unsupervised and does not need labels, which is the same as our method. Since the labels are used as extra knowledge in the other four supervised attacks, they are expected to have a better performance compared with unsupervised ones. But we will show that CLGA has comparable performance and can even outperform some of the supervised baselines in some cases.



\subsubsection{Experimental Settings}
For Metattack, MinMax and PGD, we use a 2-layer GCN as the surrogate model to acquire the poisoned graphs. For all the attack methods, we first generate the poisoned graphs and then feed them to the state-of-the-art contrastive learning framework GCA proposed by Zhu et al. \cite{zhu2020graph}. Because a 2-layer GCN is used as the encoder in GCA and PolBlog does not have features, we randomly initialize a 32-dimensional feature for each node in PolBlog. We set the augmentation threshold $\sigma$ to be 1\%/5\%/10\% of the number of edges, i.e. we are allowed to modify at most 1\%/5\%/10\% edges. For instance, if the number of edges in the clean graph is 1000, then under 1\% threshold, we can add or delete at most 10 edges. For a fair comparison, we fix the hyperparameters of GCA across experiments as suggested by Zhu et al. \cite{zhu2020graph}, including the temperature $\tau$ and stochastic augmentation rates, where $\tau=0.4$ and the edge dropping rates are 0.3 and 0.4 and the feature dropping rates are 0.1 and 0.0 for two views respectively. We use Adam optimizer and set the learning rate to be 0.01.

For the downstream node classification task, we train a simple logistic regression model using the learned embeddings and report the classification accuracy. When training, we use the public split from \cite{yang2016revisiting} for Cora and CiteSeer. For PolBlog, we randomly split the nodes into 10\%/10\%/80\% train/test/val set. We run each experiment 10 times and report the average.

For downstream link prediction task, we use a 2-layer MLP as the projection head to project the learned embeddings into a new latent space. We train the MLP with negative sampling and margin loss, i.e.,
\begin{equation}
\label{eq:margin loss}
    l=-\sum_{i,j,k}\text{log}(\sigma(\beta(e_{i},e_{j})-\beta(e_{i},e_{k})))
\end{equation}
where $e_{i}$ is the projected embedding of node $i$, $\beta$ is the cosine similarity function, $\sigma$ is the sigmoid function. Node $i$ and node $j$ is a positive pair, e.g. they are linked by an observed edge. Node $i$ and node $k$ is a randomly sampled negative pair, e.g. they are not linked in the training set. For all three datasets, we split the edges into 70\%/20\%/10\% train/test/val set, and only the edges in the training set are used to train the contrastive model. We report the area under curve (AUC) score. We also run each experiment 10 times and report the average.

\begin{table*}
\begin{center}
 \begin{tabular}{c|c|c|c|c|c|c||c|c|c|c|c|c} 
\multirow{3}{*}{Attack} & \multicolumn{6}{c||}{Node Classification} & \multicolumn{6}{c}{Link Prediction}\\
\cline{2-13}
  & \multicolumn{3}{c|}{DeepWalk} & \multicolumn{3}{c||}{GCN}& \multicolumn{3}{c|}{DeepWalk} & \multicolumn{3}{c}{GCN}\\
 \cline{2-13}
  & 1\% & 5\% & 10\% & 1\% & 5\% & 10\% & 1\% & 5\% & 10\% & 1\% & 5\% & 10\% \\
 \hline
 Metattack & \underline{0.6766} & \textbf{0.6370} & \textbf{0.5554} & \textbf{0.7830} & \textbf{0.7300} & \textbf{0.6330} & \textbf{0.8777} & \underline{0.8629} & \textbf{0.8319} & \textbf{0.8655} & \textbf{0.8378} & \underline{0.8182}\\
 \cline{1-13}
 Bojchevski et al. \cite{bojchevski2019adversarial} & 0.6826 & \underline{0.6436} & \underline{0.6276} & 0.8240 & 0.7900 & 0.7950& 0.8852 & 0.8873 & 0.8810 & 0.8675 & 0.8634 & 0.8542\\
 \cline{1-13}
 \textbf{CLGA} & \textbf{0.6720} & 0.6459 & 0.6376 & \underline{0.7860} & \underline{0.7770} & \underline{0.7690} & \underline{0.8799} & \textbf{0.8612} & \underline{0.8390} & \underline{0.8667} & \underline{0.8546} & \textbf{0.8146}\\
\end{tabular}
\caption{Transferability of attacks on Cora for DeepWalk and GCN. We report the classification accuracy for node classification and AUC score for link prediction. 1\%/5\%/10\% denote the maximum number of edges allowed to be augmented. The boldfaced ones are the best and the underlined ones are the second best among the three attacks.}
\label{table:transferability}
\end{center}
\vspace{-3em}
\end{table*}

\begin{figure*}[h]
\centering
    \subfloat[Clean graph]{\includegraphics[width=0.25\linewidth]{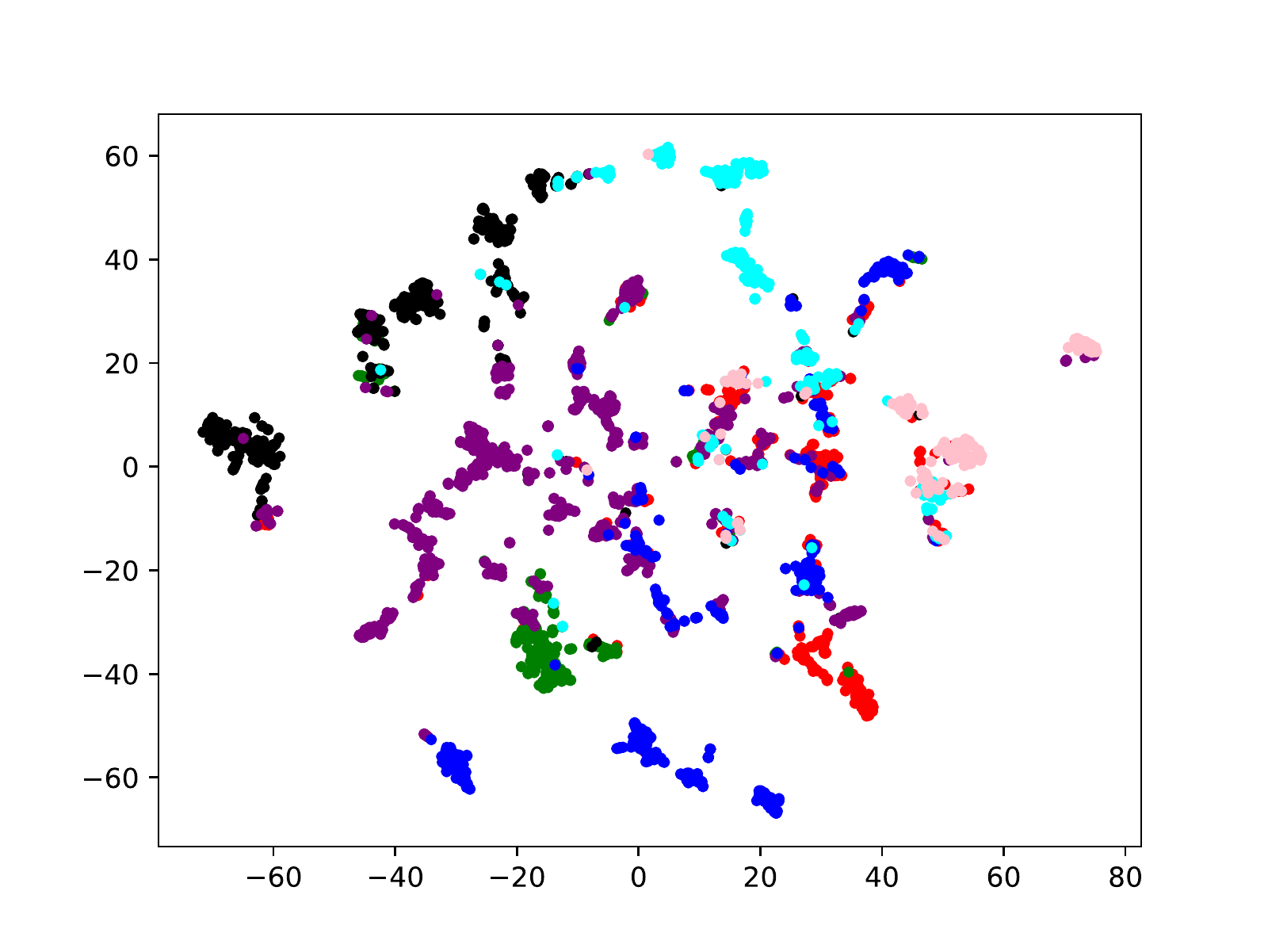}\label{fig:visual clean}}
    \subfloat[Bojchevski et al. \cite{bojchevski2019adversarial}]{\includegraphics[width=0.25\linewidth]{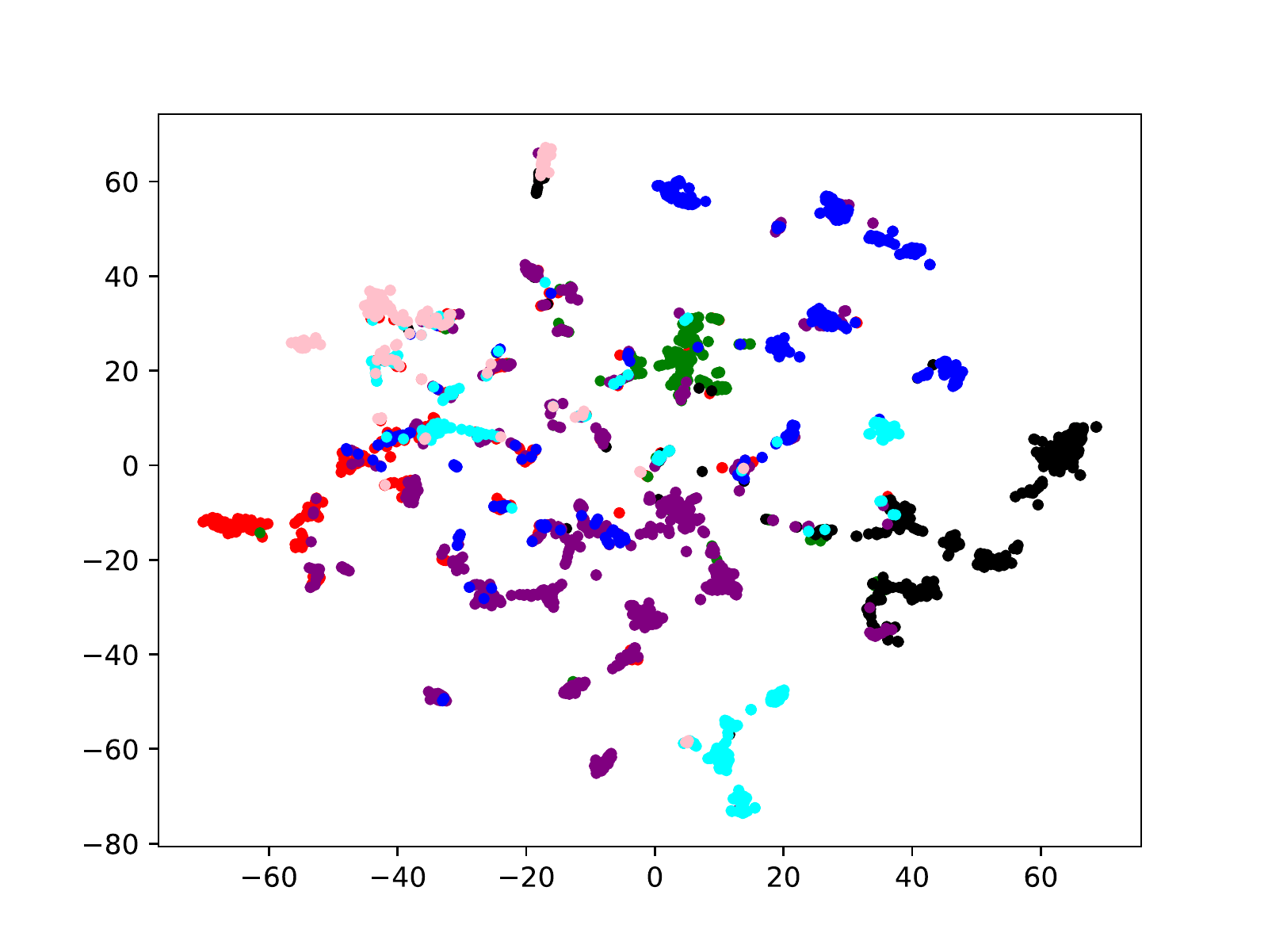}\label{fig:visual nodeembeddingattack}}
    \subfloat[Metattack]{\includegraphics[width=0.25\linewidth]{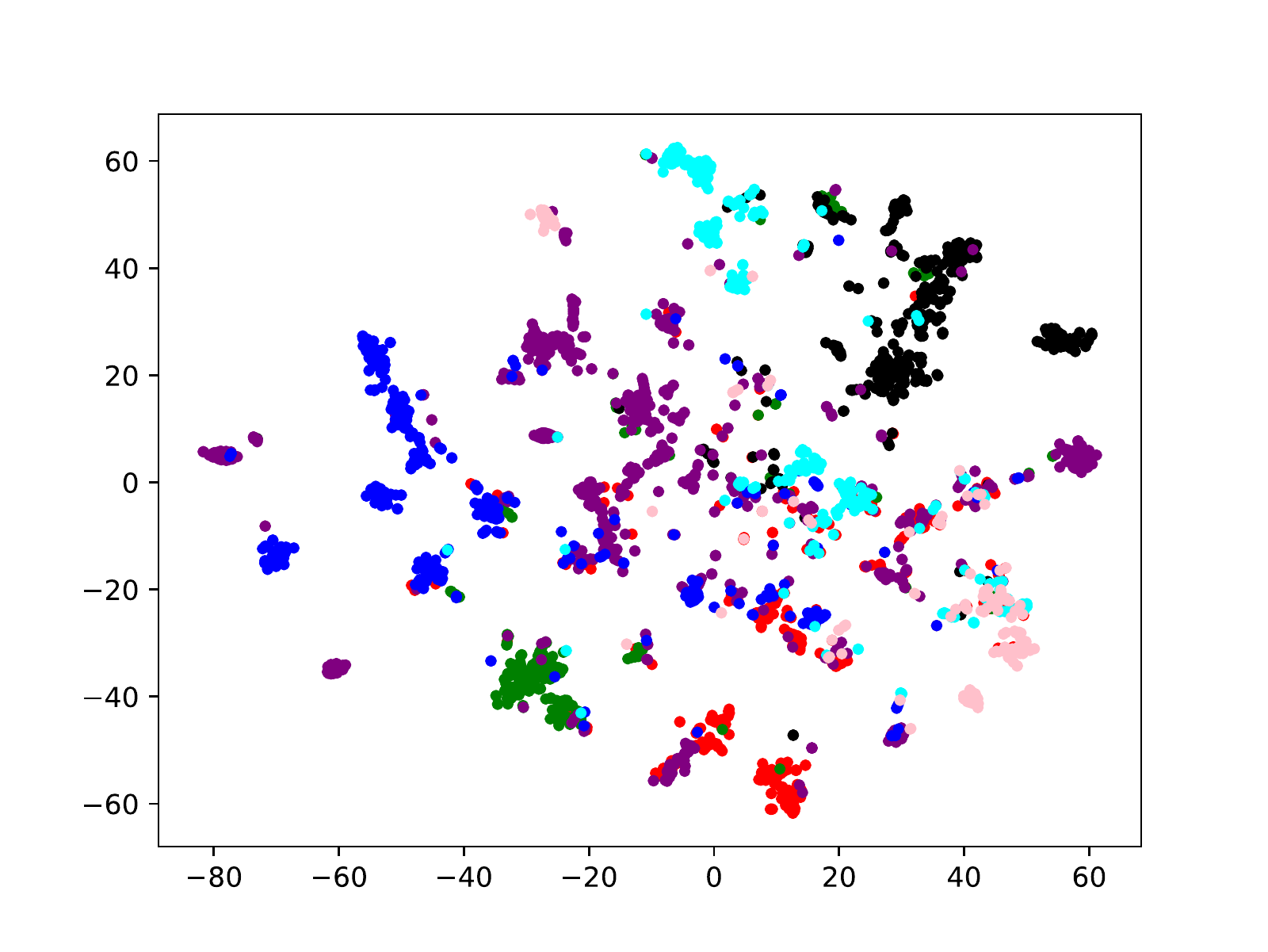}\label{fig:visual metattack}}
    \subfloat[CLGA]{\includegraphics[width=0.25\linewidth]{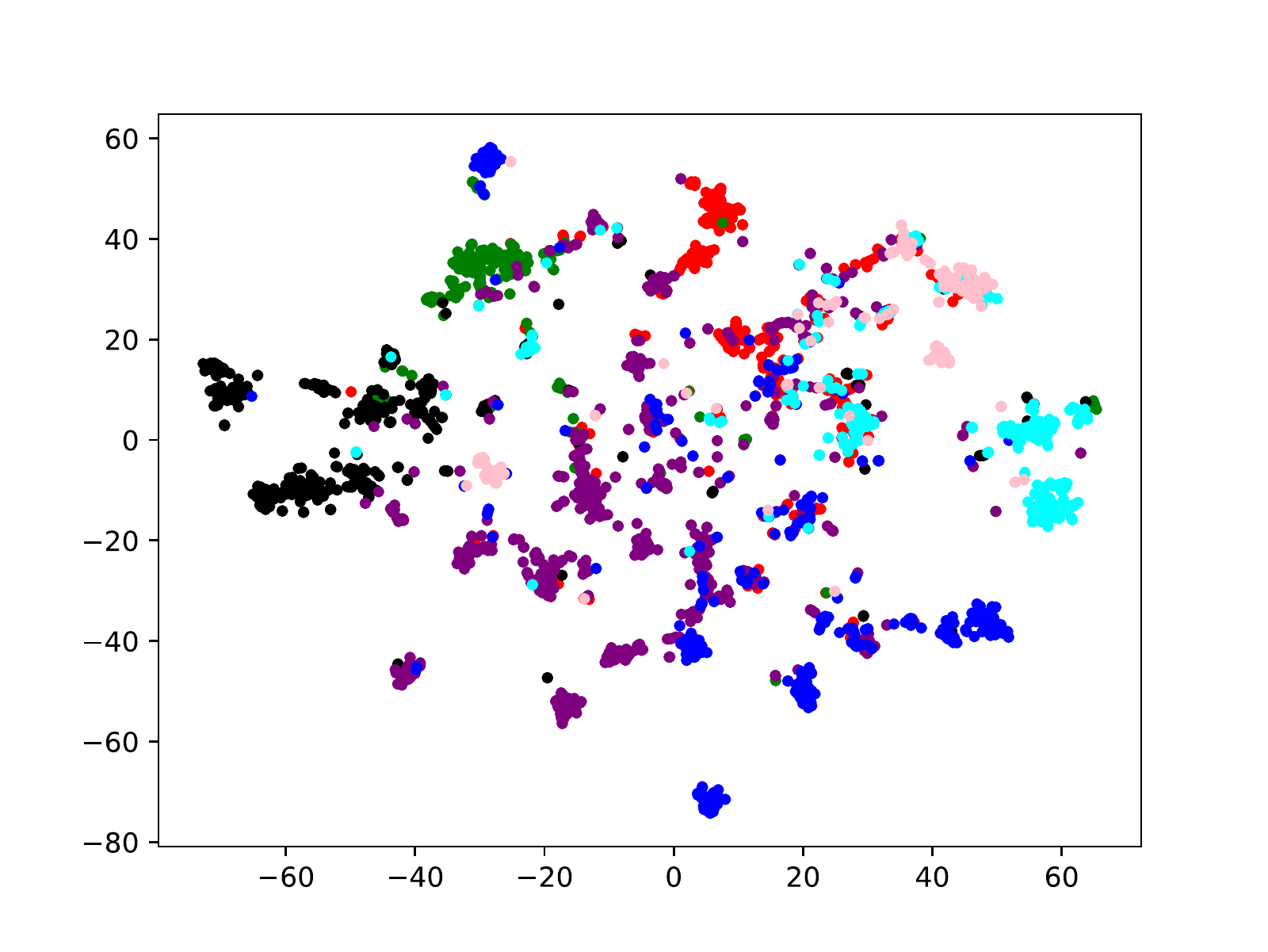}\label{fig:visual clga}}
\caption{Visualization of node embeddings learned by GCA under different attacks on Cora. Different Colors represent different classes of nodes.}
\label{fig:visualization}
\vspace{-1em}
\end{figure*}

\subsection{Comparing Node Classification Performance}
\autoref{table:node classification performance} shows the accuracy of the downstream logistic regression model trained on the embeddings learned by GCA under each attack. We can observe that, CLGA significantly outperforms the unsupervised baseline Bojchevski et al. \cite{bojchevski2019adversarial}, and can even outperform some of the supervised baselines in certain situations. For example, on Cora, CLGA outperforms PGD and DICE in all cases, and outperforms Metattack and MinMax at 1\% and 1\%/5\% augmentation rates respectively.

As the augmentation rate increases from 1\% to 10\%, the node classification accuracy continuously drops under Metattack, DICE, and CLGA. However, for the other three attacks, a larger augmentation rate does not guarantee a lower node classification accuracy. This suggests that PGD, MinMax, and Bojchevski et al. \cite{bojchevski2019adversarial} fail to capture the edges that are truly important to graph contrastive learning framework, or in other words, graph contrastive learning is robust to such attacks.

We can also observe that Metattack has the best overall performance and is very effective on Cora and CiteSeer. This is because that Metattack is originally designed for attacking an end-to-end GCN model for node classification. By utilizing the labels and the gradients, it can accurately locate the most informative edges that affect the node classification accuracy. However, such an approach is not guaranteed to work well on other downstream tasks, such as link prediction. Instead, we will show that our CLGA has a good performance not only on node prediction, but also on link prediction.

\subsection{Comparing Link Prediction Performance}
\autoref{table:link prediction performance} shows the AUC scores of the three datasets under different attacks.
We have several observations and conclusions regarding the results.

First, CLGA outperforms Bojchevski et al. \cite{bojchevski2019adversarial} except for 5\% augmentation rate on PolBlog, showing that CLGA is more effective than Bojchevski et al. \cite{bojchevski2019adversarial}. Second, CLGA has comparable performance with supervised attacks. In most cases, CLGA is the second best among all six attacks. This suggests that, even without labels, CLGA can effectively reduce the downstream link prediction performance for graph contrastive learning and achieve comparable or even better performance than supervised attacks.

Together with the observations in the node classification scenario, our unsupervised attack CLGA is shown to be able to reduce both the downstream node classification performance and link prediction performance of graph contrastive learning, and has comparable performance with state-of-the-art supervised untargeted poisoning attacks.

\subsection{Transferability}
Although CLGA has a good performance on graph contrastive learning, it is still unknown how it works for other graph representation models. Here we further compare the transferability of CLGA with Bojchevski et al. \cite{bojchevski2019adversarial} and Metattack. We use the poisoned graphs generated by the three attacks to train DeepWalk and a 2-layer GCN. We evaluate the quality of the embeddings on both node classification and link prediction. For node classification, the learned embeddings are used to train a logistic regression model. For link prediction, we use a 2-layer MLP as a projection head and train it with the loss in \autoref{eq:margin loss}. We report the accuracy for node classification and AUC score for link prediction in \autoref{table:transferability}. Other experimental settings are kept consistent with previous experiments. Note that even though the attack proposed by Bojchevski et al. \cite{bojchevski2019adversarial} is originally designed for methods based on random walks such as DeepWalk, our CLGA can still achieve comparable performance with it on DeepWalk for the node classification task, and even outperforms it on the link prediction task. Moreover, CLGA outperforms Bojchevski et al. \cite{bojchevski2019adversarial} on GCN. And CLGA even has comparable performance with Metattack at 1\% augmentation rate for node classification, and at 1\%/10\% augmentation rates for link prediction. These observations suggest that CLGA can be transferred to other graph representation models.

\subsection{Visualization}
We visualize the learned embeddings by GCA under Metattack, Bojchevski et al. \cite{bojchevski2019adversarial}, and CLGA, to show how our attack influences the distribution of embeddings. The t-SNE \cite{van2008visualizing} results are shown in \autoref{fig:visualization}. We can observe that the distributions of embeddings by Metattack and CLGA are denser than Bojchevski et al. \cite{bojchevski2019adversarial} and clean embeddings, especially in the center, which is considered to be the main reason why Metattack and CLGA outperform others. Nodes in the center are hard to be well classified because such nodes have a similar distance to each cluster and we can't tell which class it belongs to with high confidence. This suggests that CLGA can reduce the quality of the embeddings by pushing embeddings to the center in \autoref{fig:visualization}, or in other words, pushing them close to the intersections/decision boundaries of the clusters, which makes the embeddings harder to be classified.


\section{Conclusion}\label{sec:con}
In this paper, we introduce Contrastive Loss Gradient Attack (CLGA), an unsupervised untargeted poisoning attack for attacking graph contrastive learning. This is the first work to attack graph contrastive learning in an unsupervised manner without using labels. The quality of the learned embeddings is damaged and the performance of various downstream tasks is degraded. We show by extensive experiments that our CLGA outperforms unsupervised baselines and has comparable and even better performance with supervised baselines. We also show that CLGA can be transferred to other graph representation models such as DeepWalk and GCN.
\section*{Acknowledgment}
The work has been supported by Australian Research Council under grants DP220103717, DP200101374, LP170100891, and LE220100078.
\balance
\bibliographystyle{ACM-Reference-Format}
\bibliography{ref.bib}
\end{document}